\documentclass[sigconf]{aamas}

\usepackage{balance} % for balancing columns on the final page

\doi{JQIR9876}

\makeatletter
\gdef\@copyrightpermission{
  \begin{minipage}{0.2\columnwidth}
   \href{https://creativecommons.org/licenses/by/4.0/}{\includegraphics[width=0.90\textwidth]{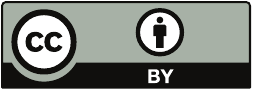}}
  \end{minipage}\hfill
  \begin{minipage}{0.8\columnwidth}
   \href{https://creativecommons.org/licenses/by/4.0/}{This work is licensed under a Creative Commons Attribution International 4.0 License.}
  \end{minipage}
  \vspace{5pt}
}
\makeatother

\setcopyright{ifaamas}
\acmConference[AAMAS '26]{Proc.\@ of the 25th International Conference
on Autonomous Agents and Multiagent Systems (AAMAS 2026)}{May 25 -- 29, 2026}
{Paphos, Cyprus}{C.~Amato, L.~Dennis, V.~Mascardi, J.~Thangarajah (eds.)}
\copyrightyear{2026}
\acmYear{2026}
\acmDOI{}
\acmPrice{}
\acmISBN{}

\acmSubmissionID{1080}

\title[AAMAS-2026 Formatting Instructions]{Human-Inspired Context-Selective Multimodal Memory for Social Robots}

\author{Hangyeol Kang}
\affiliation{
  \institution{Department of Computer Science, University of Geneva}
  \city{Geneva}
  \country{Switzerland}}
\email{hangyeol.kang@unige.ch}

\author{Slava Voloshynovskiy}
\affiliation{
  \institution{Department of Computer Science, University of Geneva}
  \city{Geneva}
  \country{Switzerland}}
\email{Svyatoslav.Voloshynovskyy@unige.ch}

\author{Nadia Magnenat Thalmann}
\affiliation{
  \institution{MIRALab, University of Geneva}
  \city{Geneva}
  \country{Switzerland}}
\email{nadia.thalmann@unige.ch}

\begin{abstract}
Memory is fundamental to social interaction, enabling humans to recall meaningful past experiences and adapt their behavior accordingly based on the context. However, most current social robots and embodied agents rely on non-selective, text-based memory, limiting their ability to support personalized, context-aware interactions. Drawing inspiration from cognitive neuroscience, we propose a context-selective, multimodal memory architecture for social robots that captures and retrieves both textual and visual episodic traces, prioritizing moments characterized by high emotional salience or scene novelty. By associating these memories with individual users, our system enables socially personalized recall and more natural, grounded dialogue. We evaluate the selective storage mechanism using a curated dataset of social scenarios, achieving a Spearman correlation of 0.506, surpassing human consistency ($\rho=0.415$) and outperforming existing image memorability models. In multimodal retrieval experiments, our fusion approach improves Recall@1 by up to 13\% over unimodal text or image retrieval. Runtime evaluations confirm that the system maintains real-time performance. Qualitative analyses further demonstrate that the proposed framework produces richer and more socially relevant responses than baseline models. This work advances memory design for social robots by bridging human-inspired selectivity and multimodal retrieval to enhance long-term, personalized human-robot interaction.
\end{abstract}


\newcommand{\BibTeX}{\rm B\kern-.05em{\sc i\kern-.025em b}\kern-.08em\TeX}
\usepackage{algorithm}
\usepackage{algorithmic}
\usepackage{multirow}

\begin{document}

%%% The following commands remove the headers in your paper. For final 
%%% papers, these will be inserted during the pagination process.

\pagestyle{fancy}
\fancyhead{}

%%% The next command prints the information defined in the preamble.

\maketitle 

%%%%%%%%%%%%%%%%%%%%%%%%%%%%%%%%%%%%%%%%%%%%%%%%%%%%%%%%%%%%%%%%%%%%%%%%

\section{Introduction}

Memory plays a central role in human cognition and social interaction. The ability to selectively store and recall past experiences enables people to sustain meaningful relationships, learn from previous encounters, and adapt flexibly to changing contexts~\cite{spreng2013examining}. For artificial agents, particularly embodied robots and assistive AI systems, equipping them with memory capabilities is essential for fostering natural, effective, and personalized interactions with humans~\cite{kwon2025embodied}.

In real-world environments, agents operate within a continuous stream of multimodal sensory information while engaging in complex social exchanges. To act as credible social partners, robots and AI need to remember not only factual events but also how and when those were meaningful to specific individuals. This requires memory systems that go beyond non-selective or text-based storage, enabling multimodal retention and retrieval of contextually meaningful experiences. For embodied agents such as social robots, contextually selective memory grounded in both perceptual and social cues is critical for developing user-aligned behavior.

Despite recent advances in conversational AI and embodied agents, most existing systems still rely on non-selective, text-based memory~\cite{spitale2025vita}. Although some frameworks have introduced visual memory modules, they often remain object-centric, interval-based, or non-selective—storing scene snapshots or generic visual information at fixed intervals regardless of contextual importance~\cite{liu2024meia,huang2024vinci}. Many also convert visual memories into textual descriptions, constraining the robot’s ability to answer detailed or flexible queries when relevant cues were never explicitly transcribed~\cite{liu2023mmhqa}. This lack of contextual selectivity limits the depth of social interaction that such agents can achieve, as object- or scene-centric memory alone cannot capture what makes an experience personally or socially meaningful. Few systems prioritize what is memorable from a human perspective, neglecting factors such as emotional salience, novelty, and user-centered relevance~\cite{peller2023memory}. As a result, they often miss opportunities to support richer, more personalized, and emotionally resonant interactions with humans.

Insights from cognitive neuroscience demonstrate that human memory is inherently selective: people do not store every moment of experience, but instead encode memories according to intrinsic and extrinsic cues~\cite{van2023turning}. Intrinsic cues, such as the perceptual complexity or distinctiveness of a scene, and extrinsic factors, such as emotional intensity, novelty, or personal relevance, are known to strongly influence which events are remembered and which are forgotten~\cite{kyle2025scene,kramer2023features,bylinskii2021memorability}. This selectivity allows people to prioritize significant, emotionally resonant, or socially important experiences, an ability that is essential for adaptive social functioning and efficient use of cognitive resources~\cite{van2023turning}.

For social robots and embodied AI, adopting a similar selective approach is essential. Real-world environments are dynamic and information-rich, and storing every sensory input or conversational exchange is neither scalable nor meaningful. Memory systems for social agents, therefore, need to integrate both selectivity and multimodality, capturing not only textual or visual information but also when an experience holds emotional or contextual significance for the user. Without such mechanisms, artificial agents risk being overwhelmed by irrelevant details and failing to form genuine, user-aligned social connections. Consequently, there is a clear need for frameworks that integrate selective and multimodal memory storage and retrieval inspired by how humans prioritize, store, and recall their most memorable experiences.

In this work, we introduce SUMMER (Selectivity Unified Multimodal Memory for Embodied Robots), an end-to-end framework that equips social robots with context-selective and multimodal memory capabilities. Grounded in principles of human memory selectivity, the framework stores and retrieves information that is not only factual but also socially and emotionally meaningful. SUMMER operates without additional model training or fine-tuning, enabling efficient integration into diverse robotic platforms. Its lightweight and modular design ensures practical deployment in real-world social and embodied agents.

To systematically evaluate the framework, we conduct four complementary studies. First, we curate a pilot dataset of social interaction scenarios to analyze how emotional salience, novelty, and scene complexity influence selective memory storage. Second, we benchmark multimodal retrieval on public datasets (Flickr8k~\cite{hodosh2013framing}, Flickr30k~\cite{young2014image}, and MS COCO~\cite{lin2014microsoft}), comparing fusion-based retrieval against unimodal text and image baselines. Third, we assess runtime performance using the pilot dataset to verify the framework’s suitability for real-time social interaction. Finally, we perform qualitative evaluations on the same dataset, comparing responses generated by the proposed framework with those of a baseline vision–language model to illustrate its ability to produce richer and more socially grounded outputs. Together, these evaluations demonstrate the framework’s capacity to capture socially meaningful experiences, retrieve contextually relevant information, and operate efficiently in dynamic, human-centered environments.

Our main technical contributions are as follows:
\begin{itemize}
    \item We present an end-to-end framework for multimodal memory storage and retrieval in social robots, integrating selective capture guided by contextual cues.
    \item We design a train-free multimodal retrieval mechanism that enables robots to respond to broader, human-like memory queries by leveraging both textual and visual modalities.
    \item We conduct comprehensive quantitative and qualitative evaluations using both public benchmarks and a new human-annotated pilot dataset focused on socially relevant, user-centered memories.
\end{itemize}

%%%%%%%%%%%%%%%%%%%%%%%%%%%%%%%%%%%%%%%%%%%%%%%%%%%%%%%%%%%%%%%%%%%%%%%%

\section{Related Work}
\subsection{Memory Systems in Social Robots}
Memory enables social robots to sustain coherent interactions, personalize behavior, and adapt to human partners over time. Early approaches focused on text-based conversational memory, using context windows or prompt-based retrieval to maintain dialogue coherence in large language models (LLMs)~\cite{mei2025survey}. Frameworks such as MemoryBank~\cite{zhong2024memorybank} and Memory Sandbox~\cite{huang2023memory} extend this concept with long-term conversational traces, autonomous recall, and user-level inspection. However, text-centric memories often lose perceptual richness when multimodal inputs are reduced to language~\cite{janssens2025multimodal}. To enable more structured reasoning, symbolic or knowledge-based systems explicitly represent events and facts using graphs or databases. Examples include ChatDB, which integrates SQL-based retrieval for multi-hop reasoning~\cite{hu2023chatdb}, and ROSA, which dynamically adapts robot behaviors based on stored contextual knowledge~\cite{rezendesilva2025rosa}.

Recent studies emphasize multimodal memory, integrating visual, auditory, and behavioral data to bridge perception and language~\cite{duncan2024survey}. The Nadine robot combines LLM reasoning with multimodal perception to retrieve user-specific episodic memories and simulate emotional states~\cite{kang2024nadinea}, while LLM-Brain unifies perception, planning, and memory through interconnected multimodal models~\cite{mai2023llm}. Other systems leverage paralinguistic and behavioral cues to infer personality traits or improve interaction quality in specific contexts such as sign language communication~\cite{li2023multimodal}, as well as socially assistive settings that rely on contextual user information~\cite{agrawal2024shelfhelp}, underscoring the importance of perceptual grounding and social sensitivity in memory design~\cite{janssens2025multimodal}.

Despite these advances, most frameworks still store information indiscriminately or at fixed intervals, lacking selectivity based on emotional salience, novelty, or social relevance~\cite{hou2024my}. Text-dominant designs further omit non-verbal and contextual cues critical for nuanced social understanding~\cite{thompson2025social}. Few systems explicitly connect memory to emotional or social context, limiting robots’ capacity to interpret atmosphere, respect norms, and adapt to user personality~\cite{kang2024nadinea}. These challenges highlight the need for more human-aligned mechanisms for selective and contextually grounded memory formation.

\subsection{Image Memorability and Selective Memory}
Beyond robotic architectures, research on image memorability offers complementary insights into how selective encoding emerges in human cognition. Large-scale studies such as LaMem~\cite{khosla2015understanding}, SUN Memorability~\cite{isola2011understanding}, and FIGRIM~\cite{bylinskii2015intrinsic} demonstrate that memorability is an intrinsic and consistent property of images across observers. Early work showed that low-level features like color or contrast are poor predictors, whereas semantic attributes—such as the presence of people or faces and overall scene structure—strongly influence recall~\cite{isola2011understanding}. Emotional salience and contextual distinctiveness further enhance memorability, suggesting that recall depends on meaningful, socially grounded features~\cite{bylinskii2021memorability}.

Deep learning models such as MemNet~\cite{khosla2015understanding}, AMNet~\cite{fajtl2018amnet}, and ViTMem~\cite{hagen2023image} now approach human-level prediction accuracy, yet remain centered on static, object- or scene-focused datasets and overlook social or interactive dynamics. From a cognitive perspective, memory is a selective process shaped by affective and contextual cues rather than a passive record~\cite{mather2007emotional}. Emotion and novelty determine which experiences are encoded and consolidated~\cite{lalumiere2017emotional,stewardson2022episodic}, while social significance further governs what is remembered~\cite{merck2020remembering}. Robotic memory systems, however, typically focus solely on visual salience~\cite{kim2024machine}, limiting their ability to capture the interactive and social nature of real-world memory.

% Our work builds on these findings by embedding cognitive principles into a selective memory mechanism for embodied agents. The proposed approach prioritizes episodic traces based on affective importance, novelty, and contextual relevance, moving beyond static prediction toward memory systems that better reflect human cognition and support socially meaningful interaction.

%%%%%%%%%%%%%%%%%%%%%%%%%%%%%%%%%%%%%%%%%%%%%%%%%%%%%%%%%%%%%%%%%%%%%%%%

\begin{figure*}[t]
    \centering
    \includegraphics[width=0.9\textwidth]{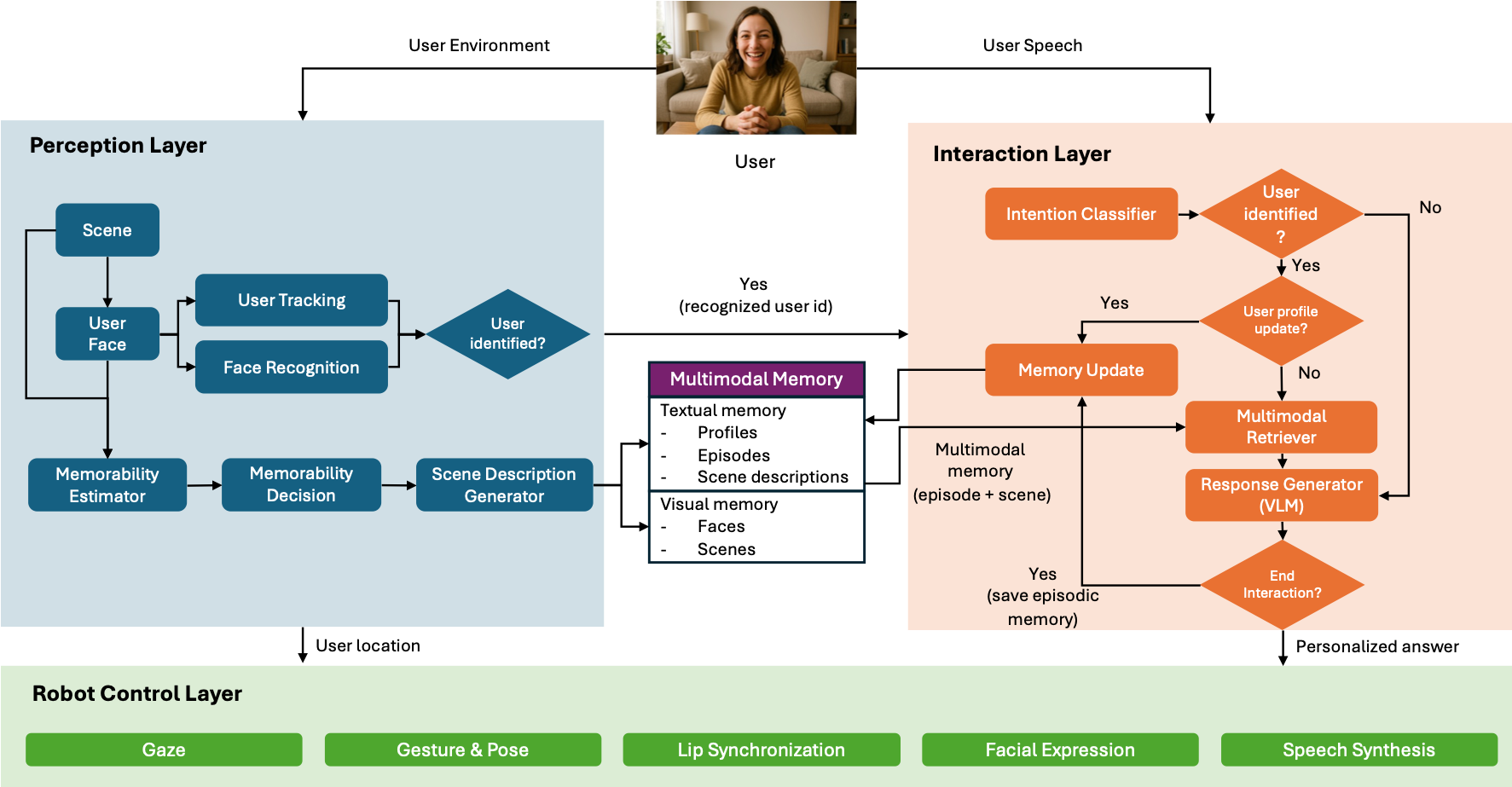} % Reduce the figure size so that it is slightly narrower than the column.
    \caption{Overview of the SUMMER architecture for selective multimodal memory in social robots. The perception layer analyzes the user's scene and face to estimate memorability; memorable scenes are stored along with generated scene descriptions. The interaction layer manages and retrieves contextually significant multimodal memories, enabling socially relevant and personalized responses.}
    \label{fig1}
    \Description{SUMMER architecture}
\end{figure*}

\section{Method}
\subsection{Overview of the SUMMER Framework}
\label{sec:3.1}

The SUMMER framework is structured into three main layers: the Perception Layer, the Interaction Layer, and the Control Layer, as depicted in Figure~\ref{fig1}. \\

\noindent\textbf{Perception Layer.}\quad
This layer analyzes the user's environment and facial expressions to estimate the memorability of each encountered scene. When a scene meets predefined selectivity criteria, it is encoded and stored in the multimodal memory database along with a textual scene description automatically generated by a lightweight vision-language model. These descriptions later facilitate efficient and contextually relevant retrieval during user interactions. The detailed mechanism for selective memory storage is discussed in Section~\ref{sec:3.2}. \\

\noindent\textbf{Interaction Layer.}\quad
The interaction layer manages context-aware retrieval and episodic memory updates. Leveraging the reasoning capabilities of LLMs, the system first classifies the user's intention based on the incoming query, for example, updating a user profile, ending a conversation, or continuing the dialogue. Memory-related modules (memory update and memory retrieval) are activated once the user is identified. Upon identification, user-specific information is retrieved from the multimodal memory database, enabling the response generation module to produce more personalized and contextually grounded replies. Details on intention classification, user identification, and multimodal retrieval processes are provided in Section~\ref{sec:3.3} and ~\ref{sec:3.4}. \\

\noindent\textbf{Control Layer.}\quad
The control layer translates high-level interaction outcomes into embodied robot behaviors, such as gaze direction, gestures, and spoken responses, grounded in the context provided by the perception and interaction layers. While not the primary focus of this work, this layer enables the social robot to deliver personalized and socially intelligent responses.

\subsection{Selective Memory Storage}
\label{sec:3.2}

Human memory is inherently selective, shaped by both intrinsic and extrinsic factors that determine which experiences are encoded and retained~\cite{xu2021predicting}. Inspired by this cognitive principle, our framework implements a context-selective storage mechanism driven by two complementary cues: emotional salience and scene novelty. Although we initially explored scene complexity as a potential intrinsic cue, empirical results showed that it did not contribute meaningfully to the selection of memorable moments in socially interactive scenarios. We therefore exclude it from the final system, focusing instead on the two cues that most robustly predict socially relevant memorability. \\

\noindent\textbf{Emotion Estimation.}\quad
Emotional salience plays a central role in memory formation, particularly in social contexts~\cite{stewardson2022episodic}. To capture this signal, we integrate a face detection and emotion recognition module that analyzes the most prominent face in each frame. (assumed to correspond to the primary user). For the detected face, we estimate both the discrete emotion category $\mathcal{E}$ = {neutral, happy, sad, surprise, fear, disgust, anger, contempt} and the corresponding intensity score $p_k$ for each emotion $k \in \mathcal{E}$.

Recognizing that not all emotions contribute equally to memorability~\cite{khosla2015understanding}, we adopt emotion-specific intensity thresholds $T_{e,k}$ that are empirically optimized through nested cross-validation. This approach allows the system to weight emotional categories according to their distinct contributions to memorability, rather than applying a uniform threshold across all emotions. It also accounts for cognitive findings that negative or high-arousal emotions tend to facilitate memory encoding more strongly than positive or low-arousal states~\cite{khosla2015understanding,marchewka2016arousal}.

For each emotion $k$, the emotion-specific salience score $s_k$ is defined as the normalized activation above its threshold:
\begin{eqnarray}
    s_k = \max\left(0, \frac{p_k - T_{e,k}}{1 - T_{e,k}}\right)
\end{eqnarray}
and the overall emotional salience of a frame is then determined by the maximum salience across all emotion categories:
\begin{eqnarray}
    e = \max_{k} \{ s_{k} \mid k \in \mathcal{E} \}
\end{eqnarray}
A frame satisfies the emotional salience condition if $e > 0$, i.e., if the intensity of at least one emotion exceeds its category-specific threshold. This formulation ensures that even a non-dominant but highly salient emotion can trigger memory encoding. The resulting optimized thresholds and their implications for selective formation are analyzed in detail in Section~\ref{sec:4}. \\

\noindent\textbf{Novelty Estimation.}\quad
Novelty serves as a complementary intrinsic cue, reflecting the distinctiveness of the current scene relative to previously encountered scenes. To quantify this, we represent each scene as an embedding vector $f_{scene}$ extracted from a vision encoder. We then compute the cosine distance between the current scene and all previously stored scenes $f_{\text{scene}_i}$, and use the minimum distance as the novelty score:
\begin{eqnarray}
    n = \min_{i} \; \text{dist}(f_{\text{scene}}, f_{\text{scene}_i})
\end{eqnarray}
where $\text{dist}(\cdot, \cdot)$ denotes cosine distance. The first scene for each user is always stored as a reference, providing a meaningful baseline for subsequent comparisons.

A new scene is considered novel if its distance from all stored scenes exceeds a novelty threshold $T_n$. This threshold-based decision rule ensures that only sufficiently distinct and non-redundant events are encoded into memory, preventing the database from being populated with repetitive or trivial scenes. In this way, the novelty mechanism complements the emotional salience module by capturing contextually meaningful moments that differ substantially from prior observations. \\

\noindent\textbf{Selective Memory Capture.}\quad
A scene is marked as memorable if it meets either the emotional salience condition or the novelty condition. For each selected moment, a lightweight vision-language model generates a short textual caption describing the event. Both the raw image and its caption, as well as their corresponding visual and textual embeddings, are stored in the multimodal memory database. These stored traces support contextually grounded retrieval and enable the system to reference past experiences in future interactions:
\begin{eqnarray}
    \text{Memorable} =
        \begin{cases}
        1, & \text{if } e > 0 \text{ or } n > T_{n} \\
        0, & \text{otherwise}
        \end{cases}
\end{eqnarray}

\begin{algorithm}[tb]
    \caption{Multimodal Hybrid Memory Retrieval in SUMMER}
    \label{alg:multimodal-retrieval}
    \footnotesize
    \begin{flushleft}
    \textbf{Input}: Query $q$; precomputed episode features $\{v_i^\text{text}\}$, scene features $\{v_j^\text{mm}\}$, and scene description features $\{d_j^\text{text}\}$; episode timestamps $\{t_i^e\}$; scene timestamps $\{t_j^s\}$\\
    \textbf{Parameter}: Text encoder $E_\text{text}$, multimodal encoder $E_\text{mm}$, scene similarity weight $\alpha \in [0,1]$, small $\varepsilon>0$\\
    \textbf{Output}: Retrieved episode $e^*$ and scene $s^*$
    \end{flushleft}
    \vspace{-2pt}
    \begin{algorithmic}[1]
        \STATE Encode query: $v_q^\text{text} = E_\text{text}(q)$,\quad $v_q^\text{mm} = E_\text{mm}(q)$
        \STATE Compute episode similarities: $S^\text{ep}_i = \mathrm{cosine}(v_q^\text{text}, v_i^\text{text})$ \; for all $i$
        \STATE Compute scene similarities:
            \begin{itemize}
                \item $S^\text{img}_j = \mathrm{cosine}(v_q^\text{mm}, v_j^\text{mm})$
                \item $S^\text{desc}_j = \mathrm{cosine}(v_q^\text{text}, d_j^\text{text})$
                \item $S^\text{scene}_j = \alpha S^\text{img}_j + (1-\alpha) S^\text{desc}_j$
            \end{itemize}
        \STATE \textit{Z-score normalize within each modality:}
        \STATE $\widehat{S}^\text{ep}_i = \dfrac{S^\text{ep}_i - \mu_\text{ep}}{\sigma_\text{ep} + \varepsilon}$ \; for all $i$ \hfill (means/SDs over $\{S^\text{ep}_i\}$)
        \STATE $\widehat{S}^\text{scene}_j = \dfrac{S^\text{scene}_j - \mu_\text{scene}}{\sigma_\text{scene} + \varepsilon}$ \; for all $j$ \hfill (means/SDs over $\{S^\text{scene}_j\}$)
        \STATE $i^* = \arg\max_i \widehat{S}^\text{ep}_i$, \quad $j^* = \arg\max_j \widehat{S}^\text{scene}_j$
        \IF{$\widehat{S}^\text{ep}_{i^*} > \widehat{S}^\text{scene}_{j^*}$}
            \STATE $e^* \leftarrow i^*$; \quad $t^* \leftarrow t^e_{i^*}$
            \STATE $s^* \leftarrow \arg\min_j \left| t^s_j - t^* \right|$
        \ELSE
            \STATE $s^* \leftarrow j^*$; \quad $t^* \leftarrow t^s_{j^*}$
            \STATE $e^* \leftarrow \arg\min_i \left| t^e_i - t^* \right|$
        \ENDIF
        \STATE \textbf{return} $e^*$, $s^*$
    \end{algorithmic}
    \hrulefill
\end{algorithm}

\subsection{Multimodal Hybrid Memory Retrieval}
\label{sec:3.3}

To support flexible and socially intelligent interaction, the SUMMER framework incorporates a hybrid retrieval mechanism that leverages both textual (conversation-based) and visual (scene-based) memories. This design enables the system to address a wide range of user queries, including references to past events, visual details, or social contexts.

When a user query is received, the system encodes the input using both a dedicated text encoder and a multimodal encoder, generating embeddings tailored to each memory modality. For textual retrieval, the query embedding generated by the text encoder is compared via cosine similarity to all user-specific textual conversation embeddings stored in the database. For visual retrieval, two similarity measures are computed: (1) the similarity between the query embedding from the multimodal encoder and stored scene embeddings, and (2) the similarity between the query embedding from the text encoder and textual scene descriptions. These two scores are then combined as a weighted sum to produce an overall scene similarity score.

To ensure fair comparison between modalities, the top similarity scores for both conversation and scene memories are z-score normalized within their respective pools. The system then selects the memory item (conversation or scene) with the highest normalized similarity as the primary retrieval result. The timestamp associated with this selected item is then used to retrieve the closest corresponding item from the other modality, thereby constructing a coherent and context-rich memory pair.

Finally, the retrieved textual (conversation) and visual (scene) memories are jointly processed by a vision-language model (VLM) to generate the final response. This integrated retrieval strategy enables the system to handle a wide spectrum of queries, whether users ask about past conversations, specific visual scenes, or require answers that bridge both modalities, thereby supporting more natural and human-like interactions.

\subsection{Supporting Modules}
\label{sec:3.4}
To enable seamless operation of the memory framework in real-world social robot applications, we incorporate several supporting modules beyond our core contributions. \\

\noindent\textbf{Intention Classifier.}\quad
The dialogue flow begins with an intention classifier that leverages LLM reasoning to infer the purpose of a user’s utterance, reducing computational overhead by activating only the necessary downstream modules. Each input is categorized as \textit{ProfileUpdate}, \textit{SessionEnd}, or \textit{Continue}. \textit{ProfileUpdate} is assigned when the user provides personal information (e.g., name, city, occupation, interests), triggering the memory update module. \textit{SessionEnd} indicates dialogue termination, prompting storage of the current exchange as an episodic memory. All remaining queries are labeled \textit{Continue}; if the user is identified, relevant user-specific memories are retrieved before response generation, otherwise the query is passed directly to the response generator. Identity verification is described in the following section. \\

\noindent\textbf{User Identification.}\quad
\begin{figure}[tb]
  \centering
  \includegraphics[width=0.8\linewidth]{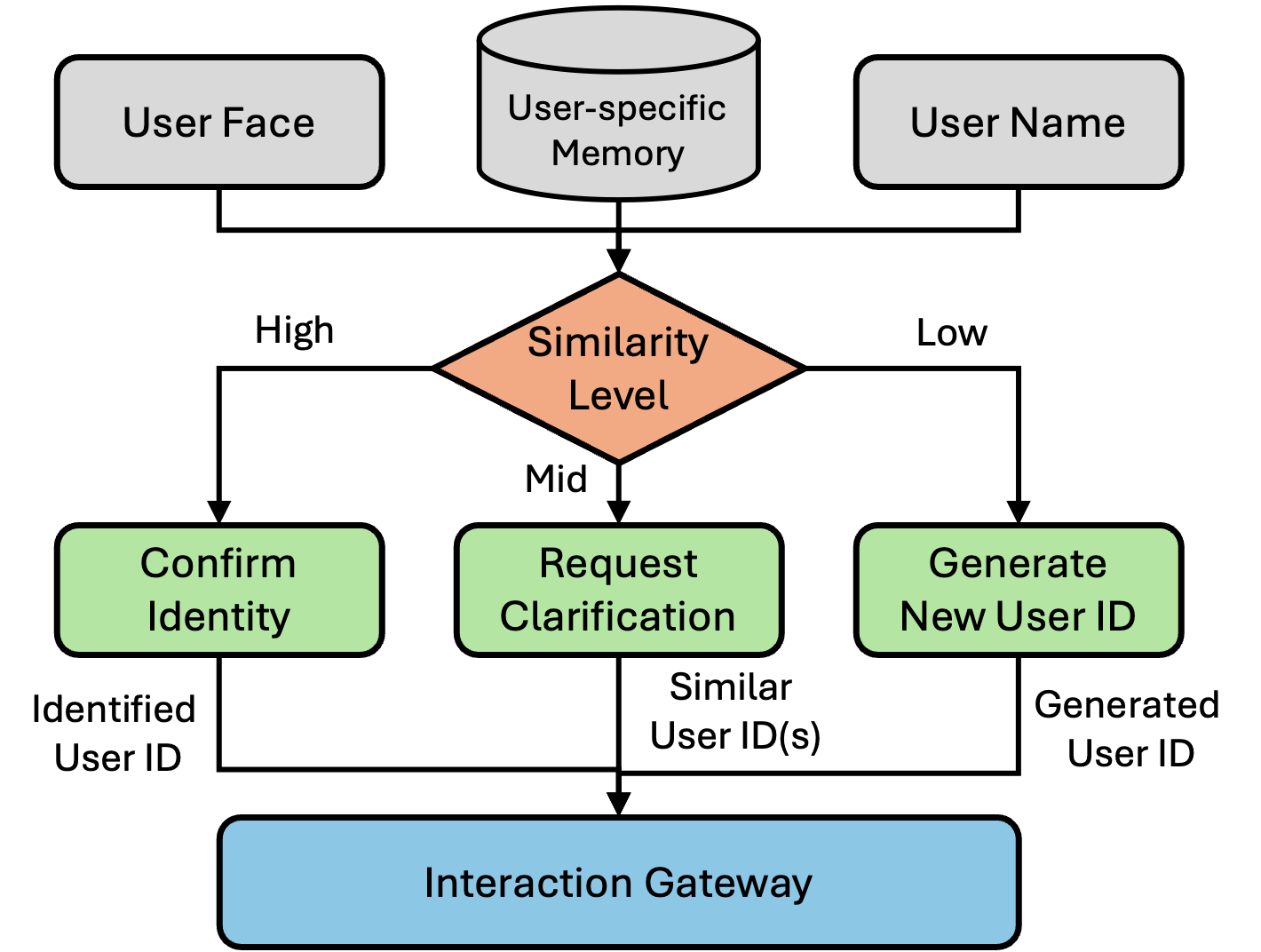}
  \caption{Overview of the user identification process}
  \label{fig2}
  \Description{Overview of the user identification process}
\end{figure}
Reliable user identification underpins consistent personalization and memory retrieval. The system verifies identity via both name and face similarity before entering the dialogue flow (Figure~\ref{fig2}). Name matching uses the Levenshtein ratio~\cite{lcvenshtcin1966binary}: scores above 0.8 confirm identity, while scores between 0.6 and 0.8 trigger a brief clarification prompt. Facial verification compares embeddings from the buffalo\_s model in InsightFace~\cite{an2021partial} using a threshold of 0.5.

If no match is found, a new user ID (e.g., 251008\_0001) is created and initialized in memory. Once identification is confirmed or established, control passes to the interaction gateway, enabling retrieval of user-specific information for personalized responses.

To protect privacy, memory operations are activated only after users voluntarily provide their name, and all stored profiles or episodic records can be permanently deleted upon request, ensuring transparency and user trust.

%%%%%%%%%%%%%%%%%%%%%%%%%%%%%%%%%%%%%%%%%%%%%%%%%%%%%%%%%%%%%%%%%%%%%%%%

\section{Experiments}
\label{sec:4}

\subsection{Experimental Setup}
\label{sec:4.1}
All experiments are conducted on 2$\times$GeForce RTX 4090 GPUs using the PyTorch deep learning framework. The baseline VLM for response generation is Mistral-small3.2~\cite{mistralai2024mistral}. For emotion detection, we employ OpenFace3.0~\cite{hu2025openface}. Novelty is assessed with CLIP-ViT-L14~\cite{radford2021learning}, and scene complexity is estimated with ICNet~\cite{feng2022ic9600}. Text features are extracted with the jina-embeddings-v4 model~\cite{gunther2025jina}, while visual encoding utilizes SigLip-so400m-P14~\cite{zhai2023sigmoid}. Scene descriptions for memorable frames are generated by Moondream2~\cite{moondream2}. All feature embeddings and multimodal memory storage are managed using ChromaDB as the vector database.

\subsection{Selective Memory Storage Evaluation}

\noindent\textbf{Curated Pilot Dataset.}\quad
To specifically assess our selective memory storage module, focusing on the memorability of social interactions and contextually significant moments, we constructed a custom pilot dataset. This dataset contains 81 images depicting a wide range of social scenarios, systematically generated using Sora, OpenAI's text-to-video generative model~\cite{brooks2024video}. Images were designed to vary in user demographics, emotional expressions, and background contexts to reflect real-world diversity. For ground truth annotation, we recruited 25 human raters, who each evaluated all images for memorability on a 1-9 Likert scale. To minimize order and recency effects, images were shown in randomized order, each for 1 second. This duration follows standard memorability protocols used in prior work, which measure rapid scene encoding rather than detailed inspection~\cite{khosla2015understanding,isola2011understanding}. This dataset was also employed in our runtime analysis and for qualitative evaluation of our SUMMER framework.

We deliberately adopted a synthetic dataset because no existing resource captures the socially grounded and interaction-centric scenes needed to assess selective memory in social robots. Synthetic generation enabled precise control over key factors such as emotional salience, novelty, and scene complexity, which are difficult to manipulate systematically in real-world data. Although synthetic images may differ from natural distributions, their use is an established practice across fields, including medical research~\cite{el2020evaluating}, visual question answering~\cite{kim2025visual}, and autonomous driving~\cite{song2023synthetic}, where they enable controlled evaluation and reproducibility. This pilot dataset was also used in our runtime analysis and qualitative evaluation of the SUMMER framework. \\

\noindent\textbf{Human Consistency.}\quad
To contextualize the performance of our model, we estimated human consistency by computing the average Spearman correlation~\cite{spearman1904correlation} between each participant's ratings and the mean rating of all other annotators. This yielded a human consistency score of $\rho$ = 0.4152, which reflects the reliability of human judgments on our dataset and serves as a practical reference point for model evaluation. Although our model achieves a higher correlation with the aggregated human ratings than this value, this does not imply that it surpasses human memory performance. Instead, it indicates that the human consistency score represents a baseline indicator of inter-annotator agreement rather than a strict upper bound. \\

\noindent\textbf{Baselines.}\quad
We compare our framework against heuristic baselines and state-of-the-art image memorability models. The heuristic baselines include random selection (uniform scores in [0,1]) and interval-based selection, which marks frames as memorable at fixed temporal intervals (e.g., every fifth or tenth frame). Both were evaluated using the same 5-fold stratified outer cross-validation with 20 repeats as our main experiments. We also evaluate two representative memorability models, ResMem~\cite{needell2022embracing} and ViTMem~\cite{hagen2023image}, which predict intrinsic memorability from visual features. For each test image, we compute predicted scores and measure Spearman correlation with human ratings using the same protocol. Together, these baselines provide heuristic and intrinsic memorability references for comparison with our approach. \\

\noindent\textbf{Model Output.}\quad
For each image in the evaluation set, our model produced a continuous memorability score based on the weighted combination of emotional salience, scene novelty, and visual complexity. Emotion probabilities were obtained using the OpenFace model~\cite{hu2025openface} applied to the largest detected face in the scene, and visual complexity was estimated with the pretrained model described in the Experimental Setup.

Novelty estimation was adapted for the evaluation phase to reflect comparative distinctiveness better. Unlike the deployment setting, where novelty is defined as the embedding distance between the current scene and previously stored ones, we employed a repeated burn-in approach to mitigate inflated scores for early images. In each run, the first $k=5$ images were excluded from novelty computation. For every subsequent image, we computed the minimum cosine distance to all preceding images in that run. This run was repeated 1000 times with randomly shuffled image sequences, and the final novelty score for each image was obtained by averaging across all repetitions. These novelty scores were then used in combination with a learned novelty threshold $T_n$ to determine whether a scene was sufficiently distinct.

The overall memorability score was computed as:
\begin{eqnarray}
\text{MemScore} & = & w_e \cdot S_{e} + w_n \cdot S_{n} + w_c \cdot S_{c}
\end{eqnarray}
where $S_{e}$, $S_{n}$, and $S_{c}$ are the emotion, novelty, and complexity scores, respectively, and $w_e$, $w_n$, $w_c$ are their respective weights. 

We evaluated the model using repeated stratified nested cross-validation (5 outer folds, 3 inner folds, 20 repeats). For each weight configuration, the inner loop searched for the optimal per-emotion thresholds and novelty threshold that maximized Spearman correlation with human ratings. These optimal parameters were then applied to the outer test folds to generate memorability predictions, which were compared against human scores using fold-level Spearman correlations and Fisher-combined $p$-values~\cite{fisher1970statistical}. \\

\begin{table}[t]
  \centering
  \caption{Spearman correlation ($\rho$), standard deviation, and significance level ($p_{\mathrm{Fisher}}$) for human consistency, baseline models, and SUMMER variants. 
  For SUMMER, the triplet denotes $(w_{\mathrm{emotion}},\,w_{\mathrm{novelty}},\,w_{\mathrm{complexity}})$ in the memorability score.
  Significance: ** $p<10^{-100}$, * $p<10^{-10}$, ns = not significant. Bold indicates the best-performing configuration.}
  \label{tab:corr_scores}
  \begin{tabular}{@{}llccc@{}}
    \toprule
    \textbf{Category} & \textbf{Approach} & \textbf{Mean $\rho$} & \textbf{Std $\rho$} & \textbf{$p_{\mathrm{Fisher}}$} \\
    \midrule
    Human & Consistency & 0.415 & — & — \\
    \midrule
    \multirow{3}{*}{Heuristic}
       & Random Selection    & -0.000  & 0.269   & ns \\
       & Interval (n=5)      & -0.052  & 0.226   & ns \\
       & Interval (n=10)     & 0.013   & 0.230   & ns \\
    \midrule
    \multirow{2}{*}{Image Mem.}
       & ResMem~\cite{needell2022embracing} & -0.055 & 0.249 & ns \\
       & ViTMem~\cite{hagen2023image}       & 0.046  & 0.235 & ns \\
    \midrule
    \multirow{5}{*}{SUMMER}
       & \textbf{(0.5,\,0.5,\,0.0)}  & \textbf{0.506} & 0.166 & \textbf{**} \\
       & (1.0,\,0.0,\,0.0)  & 0.474 & 0.170 & ** \\
       & (0.5,\,0.3,\,0.2)  & 0.447 & 0.170 & ** \\
       & (0.5,\,0.0,\,0.5)  & 0.319 & 0.188 & * \\
       & (0.0,\,1.0,\,0.0)  & -0.014 & 0.201 & ns \\
       & (0.0,\,0.0,\,1.0)  & -0.037 & 0.249 & ns \\
    \bottomrule
  \end{tabular}
\end{table}

\noindent\textbf{Results.}\quad
Table~\ref{tab:corr_scores} summarizes the predictive performance of all evaluated models in terms of Spearman correlation with human memorability ratings and their associated statistical significance. Heuristic approaches such as random and interval-based selection exhibited near-zero or weak correlations, indicating that memorability cannot be reliably predicted through simple time-based or random sampling strategies. Image memorability models originally trained on large-scale datasets, including ResMem~\cite{needell2022embracing} and ViTMem~\cite{hagen2023image}, also failed to generalize to our socially grounded interaction scenarios, showing no significant correlation with human judgments.

Although scene complexity was initially considered as an intrinsic cue, it showed little predictive value ($\rho$ = –0.037). Novelty alone also performed poorly ($\rho$ = –0.014). In contrast, combining emotional salience with novelty improved alignment with human ratings ($\rho$ = 0.506) compared to emotion alone ($\rho$ = 0.474), while adding complexity reduced performance ($\rho$ = 0.319). These results indicate that scene complexity is not a useful predictor of memorability in social-interaction contexts.

To verify that this performance drop was not due to implementation issues, we additionally evaluated ResMem~\cite{needell2022embracing}, ViTMem~\cite{hagen2023image}, and SUMMER on the LaMem benchmark (16,810 test images). ResMem achieved $\rho$=0.8145, ViTMem $\rho$=0.7583, and SUMMER $\rho$=0.2062 (pooled Spearman), consistent with each model’s intended domain. This confirms correct integration and indicates that the low performance on our pilot dataset arises from domain mismatch rather than experimental error.

\begin{table*}[ht]
    \centering
    \renewcommand{\arraystretch}{0.58}
    \caption{Recall@K (\%) for \textit{Image}, \textit{Text}, and \textit{Fusion} (inline best $\alpha$) across datasets. Bold indicates the best per dataset.}
    \label{tab:retrieval-result}
    \begin{tabular}{@{}l l l ccc ccc ccc@{}}
    \toprule
    \multirow{2}{*}{\textbf{Image Encoder}} & \multirow{2}{*}{\textbf{Text Encoder}} & \multirow{2}{*}{\textbf{Mode}} 
    & \multicolumn{3}{c}{\textbf{Flickr8k}} 
    & \multicolumn{3}{c}{\textbf{Flickr30k}} 
    & \multicolumn{3}{c}{\textbf{MS COCO}} \\
    & & & R@1 & R@5 & R@10 & R@1 & R@5 & R@10 & R@1 & R@5 & R@10 \\
    \midrule
    % ---------------------- CLIP ----------------------
    \multirow{12}{*}{CLIP ViT-L/14~\cite{radford2021learning}} 
     & \multirow{3}{*}{mE5-large~\cite{wang2024multilingual}} 
     & Text & 49.2 & 76.8 & 85.6 & 54.6 & 80.2 & 86.6 & 28.8 & 52.2 & 63.3 \\
     &  & Image & 61.6 & 86.3 & 92.8 & 64.7 & 87.1 & 92.1 & 36.5 & 61.1 & 71.1 \\
     &  & Fusion (0.7) & \textbf{68.4} & \textbf{90.5} & \textbf{95.4} & \textbf{72.2} & \textbf{90.8} & \textbf{95.2} & \textbf{42.1} & \textbf{66.9} & \textbf{76.5} \\
    \cmidrule(lr){2-12}
     & \multirow{3}{*}{mGTE-base~\cite{zhang2024mgte}} 
     & Text & 51.9 & 78.8 & 87.2 & 55.6 & 79.8 & 86.1 & 30.9 & 54.7 & 65.2 \\
     &  & Image & 61.6 & 86.3 & 92.8 & 64.7 & 87.1 & 92.1 & 36.5 & 61.1 & 71.1 \\
     &  & Fusion (0.7) & \textbf{69.0} & \textbf{90.8} & \textbf{95.8} & \textbf{72.3} & \textbf{91.2} & \textbf{95.3} & \textbf{42.6} & \textbf{67.2} & \textbf{77.0} \\
    \cmidrule(lr){2-12}
     & \multirow{3}{*}{jina-embeddings-v4~\cite{gunther2025jina}} 
     & Text & 56.2 & 82.0 & 89.6 & 59.6 & 82.6 & 88.5 & 36.0 & 61.3 & 71.9 \\
     &  & Image & 61.6 & 86.3 & 92.8 & 64.7 & 87.1 & 92.1 & 36.5 & 61.1 & 71.1 \\
     &  & Fusion (0.7) & \textbf{68.9} & \textbf{91.3} & \textbf{95.9} & \textbf{72.6} & \textbf{91.1} & \textbf{95.3} & \textbf{43.9} & \textbf{69.0} & \textbf{78.0} \\
    \cmidrule(lr){1-12}
    % ---------------------- MetaCLIP ----------------------
    \multirow{12}{*}{MetaCLIP-B/32-400M~\cite{xu2023demystifying}} 
     & \multirow{3}{*}{mE5-large~\cite{wang2024multilingual}} 
     & Text & 49.2 & 76.8 & 85.6 & 54.6 & 80.2 & 86.6 & 28.8 & 52.2 & 63.3 \\
     &  & Image & 58.1 & 84.1 & 91.2 & 62.3 & 85.5 & 91.4 & 35.9 & 61.8 & 72.2 \\
     &  & Fusion (0.7) & \textbf{65.6} & \textbf{88.9} & \textbf{94.4} & \textbf{70.9} & \textbf{90.9} & \textbf{94.6} & \textbf{41.9} & \textbf{67.7} & \textbf{77.4} \\
    \cmidrule(lr){2-12}
     & \multirow{3}{*}{mGTE-base~\cite{zhang2024mgte}} 
     & Text & 51.9 & 78.8 & 87.2 & 55.6 & 79.8 & 86.1 & 30.9 & 54.7 & 65.2 \\
     &  & Image & 58.1 & 84.1 & 91.2 & 62.3 & 85.5 & 91.4 & 35.9 & 61.8 & 72.2 \\
     &  & Fusion (0.7) & \textbf{66.1} & \textbf{89.6} & \textbf{95.1} & \textbf{71.7} & \textbf{90.7} & \textbf{94.7} & \textbf{42.3} & \textbf{68.2} & \textbf{77.6} \\
    \cmidrule(lr){2-12}
     & \multirow{3}{*}{jina-embeddings-v4~\cite{gunther2025jina}} 
     & Text & 56.2 & 82.0 & 89.6 & 59.6 & 82.6 & 88.5 & 36.0 & 61.3 & 71.9 \\
     &  & Image & 58.1 & 84.1 & 91.2 & 62.3 & 85.5 & 91.4 & 35.9 & 61.8 & 72.2 \\
     &  & Fusion (0.5) & \textbf{67.2} & \textbf{90.1} & \textbf{95.1} & \textbf{72.0} & \textbf{90.7} & \textbf{94.3} & \textbf{44.6} & \textbf{70.0} & \textbf{79.5} \\
    \cmidrule(lr){1-12}
    % ---------------------- SigLIP ----------------------
    \multirow{12}{*}{SigLIP-so400m-384~\cite{zhai2023sigmoid}} 
     & \multirow{3}{*}{mE5-large~\cite{wang2024multilingual}} 
     & Text & 49.2 & 76.8 & 85.6 & 54.6 & 80.2 & 86.6 & 28.8 & 52.2 & 63.3 \\
     &  & Image & 66.1 & 86.9 & 92.5 & 71.7 & 89.2 & 93.3 & 41.8 & 65.4 & 74.5 \\
     &  & Fusion (0.7) & \textbf{70.9} & \textbf{90.4} & \textbf{95.2} & \textbf{76.1} & \textbf{92.5} & \textbf{95.5} & \textbf{45.6} & \textbf{69.3} & \textbf{78.0} \\
    \cmidrule(lr){2-12}
     & \multirow{3}{*}{mGTE-base~\cite{zhang2024mgte}} 
     & Text & 51.9 & 78.8 & 87.2 & 55.6 & 79.8 & 86.1 & 30.9 & 54.7 & 65.2 \\
     &  & Image & 66.1 & 86.9 & 92.5 & 71.7 & 89.2 & 93.3 & 41.8 & 65.4 & 74.5 \\
     &  & Fusion (0.7) & \textbf{71.6} & \textbf{90.7} & \textbf{95.4} & \textbf{76.4} & \textbf{92.4} & \textbf{95.8} & \textbf{46.1} & \textbf{70.2} & \textbf{78.9} \\
    \cmidrule(lr){2-12}
     & \multirow{3}{*}{jina-embeddings-v4~\cite{gunther2025jina}} 
     & Text & 56.2 & 82.0 & 89.6 & 59.6 & 82.6 & 88.5 & 36.0 & 61.3 & 71.9 \\
     &  & Image & 66.1 & 86.9 & 92.5 & 71.7 & 89.2 & 93.3 & 41.8 & 65.4 & 74.5 \\
     &  & Fusion (0.7) & \textbf{71.8} & \textbf{90.9} & \textbf{95.2} & \textbf{75.8} & \textbf{92.1} & \textbf{95.6} & \textbf{47.3} & \textbf{71.2} & \textbf{79.9} \\
    \bottomrule
    \end{tabular}
\end{table*}

In contrast, all variants of the proposed SUMMER framework demonstrated substantial improvements over baseline approaches. The best-performing configuration, which combined emotion and novelty cues with weights $(0.5,0.5,0.0)$, achieved a mean Spearman correlation of $0.506$, significantly surpassing both heuristic and pre-trained memorability baselines. Several other configurations also consistently exceeded the human consistency level, highlighting the robustness of the proposed memory encoding strategy.

\subsection{Multimodal Retrieval Evaluation}

\noindent\textbf{Datasets.}\quad
We evaluated the proposed multimodal retrieval mechanism on three standard benchmarks: Flickr8k~\cite{hodosh2013framing}, Flickr30k~\cite{young2014image}, and MS COCO~\cite{lin2014microsoft}. For each dataset, we used the standard Karpathy test splits~\cite{karpathy2015deep} and generated a concise textual description for each test image using a lightweight vision-language model. \\

\noindent\textbf{Baselines.}\quad
We compared our method against two unimodal retrieval settings: (1) Textual retrieval, which measures cosine similarity between the query embedding and the textual scene descriptions, and (2) Visual retrieval, which computes similarity between the query embedding and the image embeddings. These baselines isolate the contribution of each modality and serve as reference points for evaluating the proposed fusion method. \\

\noindent\textbf{Multimodal Fusion.}\quad
To leverage the complementary strengths of both modalities, we combine normalized similarity scores from the visual and textual representations using a weighted fusion scheme. The final score is computed as
\begin{eqnarray}
    \mathrm{sim}_{\mathrm{final}} & = & \alpha \cdot \mathrm{sim}_{\mathrm{img\_norm}} + (1-\alpha) \cdot \mathrm{sim}_{\mathrm{text\_norm}}
\end{eqnarray}
where $\alpha$ controls the relative contribution of each modality. We varied $\alpha$ from 0.0 to 1.0 in increments of 0.1 and reported the best-performing value for each image-text encoder combination. Higher $\alpha$ values emphasize visual similarity, while lower values prioritize textual alignment. \\

\noindent\textbf{Results.}\quad
Table~\ref{tab:retrieval-result} summarizes Recall@K performance across datasets. The proposed multimodal fusion consistently outperforms both unimodal baselines, with best results typically obtained around $\alpha=0.7$. This pattern holds across different image and text encoder pairs, confirming that even a simple, train-free fusion can effectively exploit the complementary strengths of visual and textual representations for robust multimodal retrieval.

\subsection{Runtime Analysis}

\begin{figure*}[ht]
    \centering
    \includegraphics[width=0.95\textwidth]{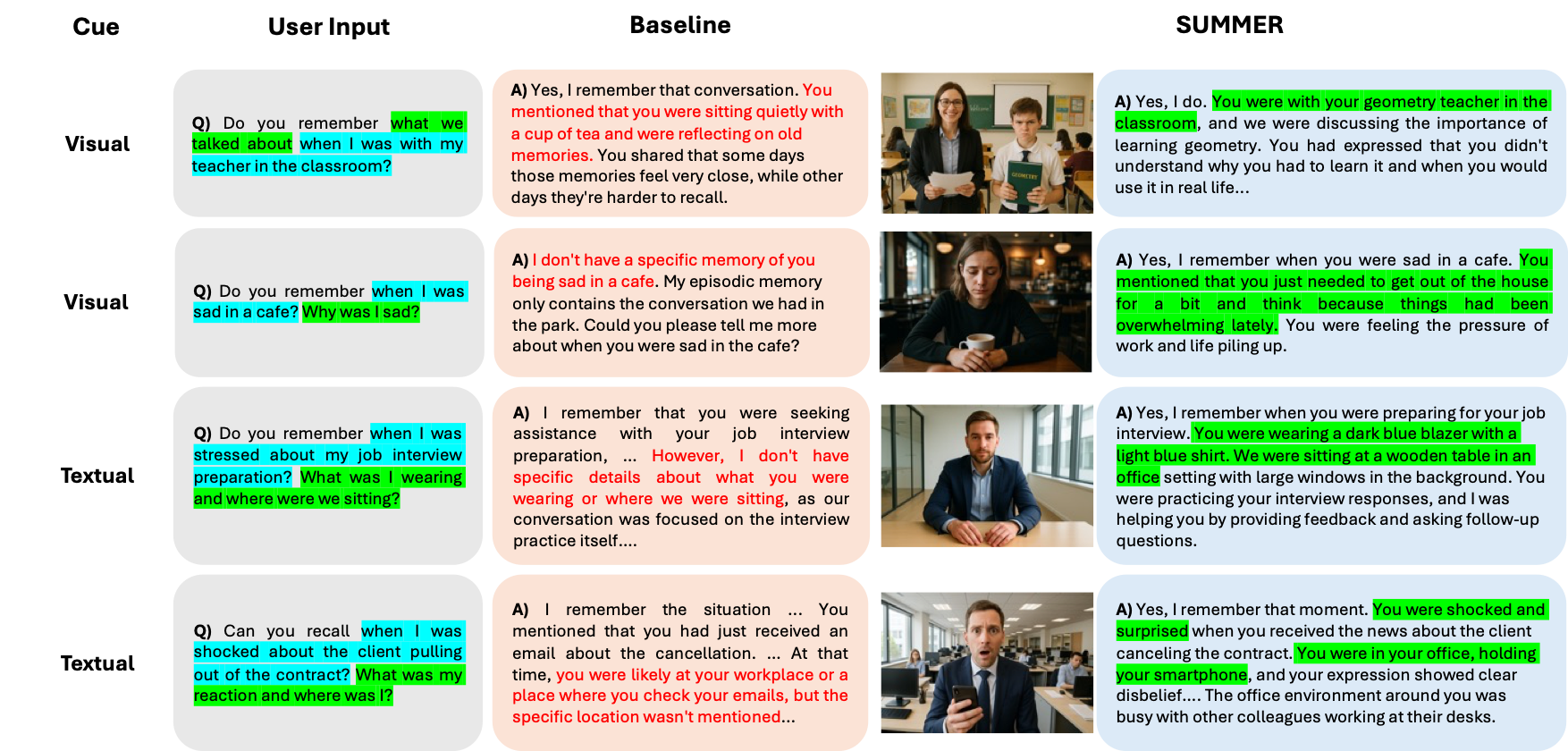}
    \caption{Qualitative comparison of response generated by the baseline model (left) and the SUMMER-augmented system (right) across diverse user queries. Each sample shows the query, both responses, and the retrieved image (for SUMMER), along with the retrieved cue type.}
    \label{fig3}
    \Description{Qualitative results}
\end{figure*}

Efficient response time is essential for maintaining natural interaction in social robots. In human–computer interaction, systems are generally expected to respond within two seconds to preserve conversational flow and prevent user frustration~\cite{shiwa2008quickly}. We evaluated the runtime performance of the baseline and SUMMER system over 100 trials, reporting the mean and standard deviation. All tests used $768\times512$ pixel inputs. The baseline employed a VLM without memory retrieval or scene analysis, while SUMMER incorporated multimodal memory operations.

As shown in Table~\ref{tab:runtime-summer}, SUMMER achieved an average generation time of $0.87\pm0.16$ seconds per response, remaining well below the two-second limit. The additional $0.4$ seconds over the baseline correspond to retrieval and image processing, enabling richer and more contextually grounded replies. The perception layer required $0.37\pm0.66$ seconds per memorable scene (Table~\ref{tab:perception-runtime}), which is acceptable since memorable events occur intermittently due to the novelty criterion. Overall, SUMMER maintains real-time performance while substantially enhancing contextual and social relevance in robot interactions.

\begin{table}[h]
  \centering
  \small
  \caption{
    Runtime of modules in the interaction layer. `Generation' refers to response generation, `Retrieval' to multimodal retrieval, and `Total Time' includes all modules in the interaction layer.
  }
  \label{tab:runtime-summer}
  \begin{tabular}{@{}l l c c@{}}
    \toprule
    System           & Module      & Time (s)         & Total Time (s)      \\
    \midrule
    Baseline         & Generation  & $0.43 \pm 0.21$  & $0.44 \pm 0.00$     \\
    \multirow{2}{*}{SUMMER}
                     & Retrieval   & $0.03 \pm 0.00$  &                     \\
                     & Generation  & $0.69 \pm 0.16$  & $0.86 \pm 0.16$     \\
    \bottomrule
  \end{tabular}
\end{table}

\begin{table}[h]
  \centering
  \small
  \caption{
    Runtime of modules in the perception layer. `Total Time' indicates the cumulative runtime for the entire perception pipeline.
  }
  \label{tab:perception-runtime}
  \begin{tabular}{@{}lcc@{}}
    \toprule
    Module                      & Time (s)           & Total Time (s)      \\
    \midrule
    Memorability Decision       & $0.08 \pm 0.01$    &                     \\
    Scene Description Generator & $0.25 \pm 0.03$    &                     \\
    Multimodal Encoding         & $0.05 \pm 0.00$    & $0.37 \pm 0.03$     \\
    \bottomrule
  \end{tabular}
\end{table}

\subsection{Qualitative Results}
To illustrate SUMMER’s advantages, we present qualitative comparisons between baseline response generation (textual memory only) and SUMMER’s multimodal approach (Figure~\ref{fig3}). When retrieval requires visual memory, such as recalling specific scenes or emotional contexts, the baseline often fails, retrieving unrelated episodes or producing incorrect answers. In contrast, SUMMER accurately recalls relevant visual memory and generates contextually grounded responses. For text-based queries, the baseline may recover the correct episode but cannot provide visually grounded details, frequently guessing or omitting visual information. By integrating both textual and visual memory, SUMMER consistently delivers more informative and socially relevant responses across both query types.

%%%%%%%%%%%%%%%%%%%%%%%%%%%%%%%%%%%%%%%%%%%%%%%%%%%%%%%%%%%%%%%%%%%%%%%%

\section{Conclusion}
In this work, we introduced SUMMER, a train-free framework for selective multimodal memory storage and retrieval for social robots. Inspired by human memory selectivity, our system enables robots to capture and recall socially and emotionally significant moments, supporting more natural and context-aware interactions. Benchmarking on public and curated datasets shows that our selective memory storage mechanism, particularly the use of emotional and novelty cues, aligns closely with human memorability. Furthermore, our multimodal retrieval module consistently outperforms standard text-only and image-only baselines. For future work, we aim to expand SUMMER’s selectivity beyond emotional salience, novelty, and scene complexity by incorporating additional factors identified in memorability research~\cite{bylinskii2021memorability}, such as social relevance, interpersonal relationships, and personalized significance.

%%%%%%%%%%%%%%%%%%%%%%%%%%%%%%%%%%%%%%%%%%%%%%%%%%%%%%%%%%%%%%%%%%%%%%%%

%%% The acknowledgments section is defined using the "acks" environment
%%% (rather than an unnumbered section). The use of this environment 
%%% ensures the proper identification of the section in the article 
%%% metadata as well as the consistent spelling of the heading.

\begin{acks}
Funded by the European Union. Views and opinions expressed are however those of the author(s) only and do not necessarily reflect those of the European Union or CINEA. Neither the European Union nor the granting authority can be held responsible for them. INDUX-R project DOI 10.3030/101135556. Funded from the Swiss State Secretariat for Education, Research and Innovation (SERI).
\end{acks}

%%%%%%%%%%%%%%%%%%%%%%%%%%%%%%%%%%%%%%%%%%%%%%%%%%%%%%%%%%%%%%%%%%%%%%%%

%%% The next two lines define, first, the bibliography style to be 
%%% applied, and, second, the bibliography file to be used.

\bibliographystyle{ACM-Reference-Format} 
\bibliography{sample}

@article{duncan2024survey,
	title        = {A {{Survey}} of {{Multimodal Perception Methods}} for {{Human}}–{{Robot Interaction}} in {{Social Environments}}},
	author       = {Duncan, John A. and Alambeigi, Farshid and Pryor, Mitchell W.},
	year         = 2024,
	month        = dec,
	journal      = {ACM Transactions on Human-Robot Interaction},
	volume       = 13,
	number       = 4,
	pages        = {1--50},
	doi          = {10.1145/3657030},
	issn         = {2573-9522},
	urldate      = {2025-10-02},
	langid       = {english}
}

@inproceedings{hou2024my,
	title        = {"{{My}} Agent Understands Me Better": {{Integrating Dynamic Human-like Memory Recall}} and {{Consolidation}} in {{LLM-Based Agents}}},
	shorttitle   = {"{{My}} Agent Understands Me Better"},
	author       = {Hou, Yuki and Tamoto, Haruki and Miyashita, Homei},
	year         = 2024,
	month        = may,
	booktitle    = {Extended {{Abstracts}} of the {{CHI Conference}} on {{Human Factors}} in {{Computing Systems}}},
	publisher    = {ACM},
	address      = {Honolulu HI USA},
	pages        = {1--7},
	doi          = {10.1145/3613905.3650839},
	isbn         = {979-8-4007-0331-7},
	urldate      = {2025-10-02},
	langid       = {english}
}

@misc{huang2023memory,
	title        = {Memory {{Sandbox}}: {{Transparent}} and {{Interactive Memory Management}} for {{Conversational Agents}}},
	shorttitle   = {Memory {{Sandbox}}},
	author       = {Huang, Ziheng and Gutierrez, Sebastian and Kamana, Hemanth and MacNeil, Stephen},
	year         = 2023,
	month        = aug,
	publisher    = {arXiv},
	number       = {arXiv:2308.01542},
	doi          = {10.48550/arXiv.2308.01542},
	urldate      = {2025-10-02},
	eprint       = {2308.01542},
	primaryclass = {cs},
	archiveprefix = {arXiv},
	langid       = {english}
}

@misc{janssens2025multimodal,
	title        = {Towards {{Multimodal Social Conversations}} with {{Robots}}: {{Using Vision-Language Models}}},
	shorttitle   = {Towards {{Multimodal Social Conversations}} with {{Robots}}},
	author       = {Janssens, Ruben and Belpaeme, Tony},
	year         = 2025,
	month        = aug,
	publisher    = {arXiv},
	number       = {arXiv:2507.19196},
	doi          = {10.48550/arXiv.2507.19196},
	urldate      = {2025-10-02},
	eprint       = {2507.19196},
	primaryclass = {cs},
	archiveprefix = {arXiv},
	langid       = {english}
}

@misc{kang2024nadinea,
	title        = {Nadine: {{An LLM-driven Intelligent Social Robot}} with {{Affective Capabilities}} and {{Human-like Memory}}},
	shorttitle   = {Nadine},
	author       = {Kang, Hangyeol and Moussa, Maher Ben and {Magnenat-Thalmann}, Nadia},
	year         = 2024,
	month        = may,
	publisher    = {arXiv},
	number       = {arXiv:2405.20189},
	doi          = {10.48550/arXiv.2405.20189},
	urldate      = {2025-10-02},
	eprint       = {2405.20189},
	primaryclass = {cs},
	archiveprefix = {arXiv},
	langid       = {english}
}

@article{li2023multimodal,
	title        = {A Multimodal Human-Robot Sign Language Interaction Framework Applied in Social Robots},
	author       = {Li, Jie and Zhong, Junpei and Wang, Ning},
	year         = 2023,
	month        = apr,
	journal      = {Frontiers in Neuroscience},
	volume       = 17,
	pages        = 1168888,
	doi          = {10.3389/fnins.2023.1168888},
	issn         = {1662-453X},
	urldate      = {2025-10-02},
	langid       = {english}
}

@article{rezendesilva2025rosa,
	title        = {{{ROSA}}: A Knowledge-Based Solution for Robot Self-Adaptation},
	shorttitle   = {{ROSA}},
	author       = {Rezende Silva, Gustavo and Päßler, Juliane and Tapia Tarifa, S. Lizeth and Johnsen, Einar Broch and Hernández Corbato, Carlos},
	year         = 2025,
	month        = may,
	journal      = {Frontiers in Robotics and AI},
	volume       = 12,
	pages        = 1531743,
	doi          = {10.3389/frobt.2025.1531743},
	issn         = {2296-9144},
	urldate      = {2025-10-02},
	langid       = {english}
}

@misc{thompson2025social,
	title        = {The {{Social Context}} of {{Human-Robot Interactions}}},
	author       = {Thompson, Sydney and Candon, Kate and Vázquez, Marynel},
	year         = 2025,
	month        = aug,
	publisher    = {arXiv},
	number       = {arXiv:2508.13982},
	doi          = {10.48550/arXiv.2508.13982},
	urldate      = {2025-10-02},
	eprint       = {2508.13982},
	primaryclass = {cs},
	archiveprefix = {arXiv},
	langid       = {english}
}

@inproceedings{zhong2024memorybank,
	title        = {Memorybank: Enhancing large language models with long-term memory},
	author       = {Zhong, Wanjun and Guo, Lianghong and Gao, Qiqi and Ye, He and Wang, Yanlin},
	year         = 2024,
	booktitle    = {Proceedings of the AAAI Conference on Artificial Intelligence},
	volume       = 38,
	pages        = {19724--19731}
}

@article{mai2023llm,
	title        = {Llm as a robotic brain: Unifying egocentric memory and control},
	author       = {Mai, Jinjie and Chen, Jun and Qian, Guocheng and Elhoseiny, Mohamed and Ghanem, Bernard and others},
	year         = 2023,
	publisher    = {arXiv}
}

@article{mei2025survey,
	title        = {A survey of context engineering for large language models},
	author       = {Mei, Lingrui and Yao, Jiayu and Ge, Yuyao and Wang, Yiwei and Bi, Baolong and Cai, Yujun and Liu, Jiazhi and Li, Mingyu and Li, Zhong-Zhi and Zhang, Duzhen and others},
	year         = 2025,
	journal      = {arXiv preprint arXiv:2507.13334}
}

@article{hu2023chatdb,
	title        = {Chatdb: Augmenting llms with databases as their symbolic memory},
	author       = {Hu, Chenxu and Fu, Jie and Du, Chenzhuang and Luo, Simian and Zhao, Junbo and Zhao, Hang},
	year         = 2023,
	journal      = {arXiv preprint arXiv:2306.03901}
}

@article{bylinskii2015intrinsic,
	title        = {Intrinsic and Extrinsic Effects on Image Memorability},
	author       = {Bylinskii, Zoya and Isola, Phillip and Bainbridge, Constance and Torralba, Antonio and Oliva, Aude},
	year         = 2015,
	month        = nov,
	journal      = {Vision Research},
	volume       = 116,
	pages        = {165--178},
	doi          = {10.1016/j.visres.2015.03.005},
	issn         = {00426989},
	urldate      = {2025-10-03},
	langid       = {english}
}

@misc{bylinskii2021memorability,
	title        = {Memorability: {{An}} Image-Computable Measure of Information Utility},
	shorttitle   = {Memorability},
	author       = {Bylinskii, Zoya and Goetschalckx, Lore and Newman, Anelise and Oliva, Aude},
	year         = 2021,
	month        = apr,
	publisher    = {arXiv},
	number       = {arXiv:2104.00805},
	doi          = {10.48550/arXiv.2104.00805},
	urldate      = {2025-10-03},
	eprint       = {2104.00805},
	primaryclass = {cs},
	archiveprefix = {arXiv},
	langid       = {english}
}

@misc{fajtl2018amnet,
	title        = {{{AMNet}}: {{Memorability Estimation}} with {{Attention}}},
	shorttitle   = {{AMNet}},
	author       = {Fajtl, Jiri and Argyriou, Vasileios and Monekosso, Dorothy and Remagnino, Paolo},
	year         = 2018,
	month        = apr,
	publisher    = {arXiv},
	number       = {arXiv:1804.03115},
	doi          = {10.48550/arXiv.1804.03115},
	urldate      = {2025-10-03},
	eprint       = {1804.03115},
	primaryclass = {cs},
	archiveprefix = {arXiv},
	langid       = {english}
}

@misc{hagen2023image,
	title        = {Image {{Memorability Prediction}} with {{Vision Transformers}}},
	author       = {Hagen, Thomas and Espeseth, Thomas},
	year         = 2023,
	month        = jan,
	publisher    = {arXiv},
	number       = {arXiv:2301.08647},
	doi          = {10.48550/arXiv.2301.08647},
	urldate      = {2025-09-28},
	eprint       = {2301.08647},
	primaryclass = {cs},
	archiveprefix = {arXiv},
	langid       = {english}
}

@article{isola2011understanding,
	title        = {Understanding the {{Intrinsic Memorability}} of {{Images}}},
	author       = {Isola, Phillip and Parikh, Devi and Torralba, Antonio and Oliva, Aude},
	year         = 2011,
	langid       = {english}
}

@inproceedings{khosla2015understanding,
	title        = {Understanding and {{Predicting Image Memorability}} at a {{Large Scale}}},
	author       = {Khosla, Aditya and Raju, Akhil S. and Torralba, Antonio and Oliva, Aude},
	year         = 2015,
	month        = dec,
	booktitle    = {2015 {{IEEE International Conference}} on {{Computer Vision}} ({{ICCV}})},
	publisher    = {IEEE},
	address      = {Santiago, Chile},
	doi          = {10.1109/iccv.2015.275},
	urldate      = {2025-07-21},
	langid       = {english}
}

@misc{kim2024machine,
	title        = {A {{Machine With Human-Like Memory Systems}}},
	author       = {Kim, Taewoon and Cochez, Michael and {Francois-Lavet}, Vincent and Neerincx, Mark and Vossen, Piek},
	year         = 2024,
	month        = aug,
	publisher    = {arXiv},
	number       = {arXiv:2204.01611},
	doi          = {10.48550/arXiv.2204.01611},
	urldate      = {2025-10-03},
	eprint       = {2204.01611},
	primaryclass = {cs},
	archiveprefix = {arXiv},
	langid       = {english}
}

@article{mather2007emotional,
	title        = {Emotional {{Arousal}} and {{Memory Binding}}: {{An Object-Based Framework}}},
	shorttitle   = {Emotional {{Arousal}} and {{Memory Binding}}},
	author       = {Mather, Mara},
	year         = 2007,
	month        = mar,
	journal      = {Perspectives on Psychological Science},
	volume       = 2,
	number       = 1,
	pages        = {33--52},
	doi          = {10.1111/j.1745-6916.2007.00028.x},
	issn         = {1745-6916, 1745-6924},
	urldate      = {2025-10-03},
	copyright    = {https://journals.sagepub.com/page/policies/text-and-data-mining-license},
	langid       = {english}
}

@misc{needell2022embracing,
	title        = {Embracing {{New Techniques}} in {{Deep Learning}} for {{Estimating Image Memorability}}},
	author       = {Needell, Coen D. and Bainbridge, Wilma A.},
	year         = 2022,
	month        = jan,
	publisher    = {arXiv},
	number       = {arXiv:2105.10598},
	doi          = {10.48550/arXiv.2105.10598},
	urldate      = {2025-09-28},
	eprint       = {2105.10598},
	primaryclass = {cs},
	archiveprefix = {arXiv},
	langid       = {english}
}

@article{xu2021predicting,
	title        = {Predicting event memorability from contextual visual semantics},
	author       = {Xu, Qianli and Fang, Fen and Molino, Ana and Subbaraju, Vigneshwaran and Lim, Joo-Hwee},
	year         = 2021,
	journal      = {Advances in Neural Information Processing Systems},
	volume       = 34,
	pages        = {22431--22442}
}

@article{marchewka2016arousal,
	title        = {Arousal rather than basic emotions influence long-term recognition memory in humans},
	author       = {Marchewka, Artur and Wypych, Marek and Moslehi, Abnoos and Riegel, Monika and Micha{\l}owski, Jaros{\l}aw M and Jednor{\'o}g, Katarzyna},
	year         = 2016,
	journal      = {Frontiers in behavioral neuroscience},
	publisher    = {Frontiers Media SA},
	volume       = 10,
	pages        = 198
}

@misc{stewardson2022episodic,
	title        = {Episodic memory through a social and emotional lens. Emotion. Advance online publication},
	author       = {Stewardson, CI and Hunsche, MC and Wardell, V and Palombo, DJ and Kerns, CM},
	year         = 2022
}

@article{lalumiere2017emotional,
	title        = {Emotional modulation of learning and memory: pharmacological implications},
	author       = {LaLumiere, Ryan T and McGaugh, James L and McIntyre, Christa K},
	year         = 2017,
	journal      = {Pharmacological reviews},
	publisher    = {Elsevier},
	volume       = 69,
	number       = 3,
	pages        = {236--255}
}

@article{merck2020remembering,
	title        = {Remembering the big game: Social identity and memory for media events},
	author       = {Merck, Clinton and Yamashiro, Jeremy K and Hirst, William},
	year         = 2020,
	journal      = {Memory},
	publisher    = {Taylor \& Francis},
	volume       = 28,
	number       = 6,
	pages        = {795--814}
}

@inproceedings{an2021partial,
	title        = {Partial fc: Training 10 million identities on a single machine},
	author       = {An, Xiang and Zhu, Xuhan and Gao, Yuan and Xiao, Yang and Zhao, Yongle and Feng, Ziyong and Wu, Lan and Qin, Bin and Zhang, Ming and Zhang, Debing and others},
	year         = 2021,
	booktitle    = {Proceedings of the IEEE/CVF International Conference on Computer Vision},
	pages        = {1445--1449}
}

@inproceedings{lcvenshtcin1966binary,
	title        = {Binary coors capable or ‘correcting deletions, insertions, and reversals},
	author       = {Lcvenshtcin, VI},
	year         = 1966,
	booktitle    = {Soviet physics-doklady},
	volume       = 10,
	number       = 8
}

@article{hu2025openface,
	title        = {OpenFace 3.0: A Lightweight Multitask System for Comprehensive Facial Behavior Analysis},
	author       = {Hu, Jiewen and Mathur, Leena and Liang, Paul Pu and Morency, Louis-Philippe},
	year         = 2025,
	journal      = {arXiv preprint arXiv:2506.02891}
}

@inproceedings{radford2021learning,
	title        = {Learning transferable visual models from natural language supervision},
	author       = {Radford, Alec and Kim, Jong Wook and Hallacy, Chris and Ramesh, Aditya and Goh, Gabriel and Agarwal, Sandhini and Sastry, Girish and Askell, Amanda and Mishkin, Pamela and Clark, Jack and others},
	year         = 2021,
	booktitle    = {International conference on machine learning},
	pages        = {8748--8763},
	organization = {PmLR}
}

@inproceedings{zhai2023sigmoid,
	title        = {Sigmoid loss for language image pre-training},
	author       = {Zhai, Xiaohua and Mustafa, Basil and Kolesnikov, Alexander and Beyer, Lucas},
	year         = 2023,
	booktitle    = {Proceedings of the IEEE/CVF international conference on computer vision},
	pages        = {11975--11986}
}

@article{feng2022ic9600,
	title        = {Ic9600: A benchmark dataset for automatic image complexity assessment},
	author       = {Feng, Tinglei and Zhai, Yingjie and Yang, Jufeng and Liang, Jie and Fan, Deng-Ping and Zhang, Jing and Shao, Ling and Tao, Dacheng},
	year         = 2022,
	journal      = {IEEE Transactions on Pattern Analysis and Machine Intelligence},
	publisher    = {IEEE},
	volume       = 45,
	number       = 7,
	pages        = {8577--8593}
}

@misc{mistralai2024mistral,
	title        = {Mistral-Small-3.2-24B-Instruct-2506},
	author       = {MistralAI},
	year         = 2025,
	publisher    = {Hugging Face},
	howpublished = {\url{https://huggingface.co/mistralai/Mistral-Small-3.2-24B-Instruct-2506}}
}

@misc{moondream2,
	title        = {Moondream2-2B},
	author       = {MoondreamLabs},
	year         = 2025,
	howpublished = {\url{https://https://huggingface.co/vikhyatk/moondream2}}
}

@article{brooks2024video,
	title        = {Video generation models as world simulators},
	author       = {Brooks, Tim and Peebles, Bill and Holmes, Connor and DePue, Will and Guo, Yufei and Jing, Li and Schnurr, David and Taylor, Joe and Luhman, Troy and Luhman, Eric and others},
	year         = 2024,
	journal      = {OpenAI Blog},
	volume       = 1,
	number       = 8,
	pages        = 1
}

@article{spearman1904correlation,
	title        = {The proof and measurement of association between two things},
	author       = {Spearman, Charles},
	year         = 1904,
	journal      = {The American Journal of Psychology},
	publisher    = {University of Illinois Press},
	volume       = 15,
	number       = 1,
	pages        = {72--101}
}

@article{young2014image,
	title        = {From image descriptions to visual denotations: New similarity metrics for semantic inference over event descriptions},
	author       = {Young, Peter and Lai, Alice and Hodosh, Micah and Hockenmaier, Julia},
	year         = 2014,
	journal      = {Transactions of the association for computational linguistics},
	publisher    = {MIT Press One Rogers Street, Cambridge, MA 02142-1209, USA journals-info~…},
	volume       = 2,
	pages        = {67--78}
}

@article{hodosh2013framing,
	title        = {Framing image description as a ranking task: Data, models and evaluation metrics},
	author       = {Hodosh, Micah and Young, Peter and Hockenmaier, Julia},
	year         = 2013,
	journal      = {Journal of Artificial Intelligence Research},
	volume       = 47,
	pages        = {853--899}
}

@inproceedings{karpathy2015deep,
	title        = {Deep visual-semantic alignments for generating image descriptions},
	author       = {Karpathy, Andrej and Fei-Fei, Li},
	year         = 2015,
	booktitle    = {Proceedings of the IEEE conference on computer vision and pattern recognition},
	pages        = {3128--3137}
}

@inproceedings{shiwa2008quickly,
	title        = {How quickly should communication robots respond?},
	author       = {Shiwa, Toshiyuki and Kanda, Takayuki and Imai, Michita and Ishiguro, Hiroshi and Hagita, Norihiro},
	year         = 2008,
	booktitle    = {Proceedings of the 3rd ACM/IEEE international conference on Human robot interaction},
	pages        = {153--160}
}

@inproceedings{lin2014microsoft,
	title        = {Microsoft coco: Common objects in context},
	author       = {Lin, Tsung-Yi and Maire, Michael and Belongie, Serge and Hays, James and Perona, Pietro and Ramanan, Deva and Doll{\'a}r, Piotr and Zitnick, C Lawrence},
	year         = 2014,
	booktitle    = {European conference on computer vision},
	pages        = {740--755},
	organization = {Springer}
}

@article{song2023synthetic,
	title        = {Synthetic datasets for autonomous driving: A survey},
	author       = {Song, Zhihang and He, Zimin and Li, Xingyu and Ma, Qiming and Ming, Ruibo and Mao, Zhiqi and Pei, Huaxin and Peng, Lihui and Hu, Jianming and Yao, Danya and others},
	year         = 2023,
	journal      = {IEEE Transactions on Intelligent Vehicles},
	publisher    = {IEEE},
	volume       = 9,
	number       = 1,
	pages        = {1847--1864}
}

@article{kim2025visual,
	title        = {Visual question answering: A survey of methods, datasets, evaluation, and challenges},
	author       = {Kim, Byeong Su and Kim, Jieun and Lee, Deokwoo and Jang, Beakcheol},
	year         = 2025,
	journal      = {ACM Computing Surveys},
	publisher    = {ACM New York, NY},
	volume       = 57,
	number       = 10,
	pages        = {1--35}
}

@article{el2020evaluating,
	title        = {Evaluating identity disclosure risk in fully synthetic health data: model development and validation},
	author       = {El Emam, Khaled and Mosquera, Lucy and Bass, Jason},
	year         = 2020,
	journal      = {Journal of medical Internet research},
	publisher    = {JMIR Publications Toronto, Canada},
	volume       = 22,
	number       = 11,
	pages        = {e23139}
}

@misc{spreng2013examining,
	title        = {Examining the role of memory in social cognition},
	author       = {Spreng, R Nathan},
	year         = 2013,
	journal      = {Frontiers in psychology},
	publisher    = {Frontiers Media SA},
	volume       = 4,
	pages        = 437
}

@article{kwon2025embodied,
	title        = {Embodied Agents Meet Personalization: Exploring Memory Utilization for Personalized Assistance},
	author       = {Kwon, Taeyoon and Choi, Dongwook and Kim, Sunghwan and Kim, Hyojun and Moon, Seungjun and Kwak, Beong-woo and Huang, Kuan-Hao and Yeo, Jinyoung},
	year         = 2025,
	journal      = {arXiv preprint arXiv:2505.16348}
}

@article{liu2024meia,
	title        = {Meia: Multimodal embodied perception and interaction in unknown environments},
	author       = {Liu, Yang and Song, Xinshuai and Jiang, Kaixuan and Chen, Weixing and Luo, Jingzhou and Li, Guanbin and Lin, Liang},
	year         = 2024,
	journal      = {arXiv preprint arXiv:2402.00290}
}

@article{liu2023mmhqa,
	title        = {MMHQA-ICL: Multimodal in-context learning for hybrid question answering over text, tables and images},
	author       = {Liu, Weihao and Lei, Fangyu and Luo, Tongxu and Lei, Jiahe and He, Shizhu and Zhao, Jun and Liu, Kang},
	year         = 2023,
	journal      = {arXiv preprint arXiv:2309.04790}
}

@article{spitale2025vita,
	title        = {VITA: A Multi-Modal LLM-Based System for Longitudinal, Autonomous and Adaptive Robotic Mental Well-Being Coaching},
	author       = {Spitale, Micol and Axelsson, Minja and Gunes, Hatice},
	year         = 2025,
	journal      = {ACM Transactions on Human-Robot Interaction},
	publisher    = {ACM New York, NY},
	volume       = 14,
	number       = 2,
	pages        = {1--28}
}

@article{van2023turning,
	title        = {Turning attention inside out: How working memory serves behavior},
	author       = {Van Ede, Freek and Nobre, Anna C},
	year         = 2023,
	journal      = {Annual review of psychology},
	publisher    = {Annual Reviews},
	volume       = 74,
	number       = 1,
	pages        = {137--165}
}

@article{kyle2025scene,
	title        = {Scene complexity and the detail trace of human long-term visual memory},
	author       = {Kyle-Davidson, Cameron and Solis, Oscar and Robinson, Stephen and Tan, Ryan Tze Wang and Evans, Karla K},
	year         = 2025,
	journal      = {Vision Research},
	publisher    = {Elsevier},
	volume       = 227,
	pages        = 108525
}

@article{kramer2023features,
	title        = {The features underlying the memorability of objects},
	author       = {Kramer, Max A and Hebart, Martin N and Baker, Chris I and Bainbridge, Wilma A},
	year         = 2023,
	journal      = {Science advances},
	publisher    = {American Association for the Advancement of Science},
	volume       = 9,
	number       = 17,
	pages        = {eadd2981}
}

@article{peller2023memory,
	title        = {A memory system of a robot cognitive architecture and its implementation in ArmarX},
	author       = {Peller-Konrad, Fabian and Kartmann, Rainer and Dreher, Christian RG and Meixner, Andre and Reister, Fabian and Grotz, Markus and Asfour, Tamim},
	year         = 2023,
	journal      = {Robotics and Autonomous Systems},
	publisher    = {Elsevier},
	volume       = 164,
	pages        = 104415
}

@article{huang2024vinci,
	title        = {Vinci: A real-time embodied smart assistant based on egocentric vision-language model},
	author       = {Huang, Yifei and Xu, Jilan and Pei, Baoqi and He, Yuping and Chen, Guo and Yang, Lijin and Chen, Xinyuan and Wang, Yaohui and Nie, Zheng and Liu, Jinyao and others},
	year         = 2024,
	journal      = {arXiv preprint arXiv:2412.21080}
}

@article{gunther2025jina,
	title        = {jina-embeddings-v4: Universal Embeddings for Multimodal Multilingual Retrieval},
	author       = {G{\"u}nther, Michael and Sturua, Saba and Akram, Mohammad Kalim and Mohr, Isabelle and Ungureanu, Andrei and Wang, Bo and Eslami, Sedigheh and Martens, Scott and Werk, Maximilian and Wang, Nan and others},
	year         = 2025,
	journal      = {arXiv preprint arXiv:2506.18902}
}

@article{zhang2024mgte,
	title        = {mgte: Generalized long-context text representation and reranking models for multilingual text retrieval},
	author       = {Zhang, Xin and Zhang, Yanzhao and Long, Dingkun and Xie, Wen and Dai, Ziqi and Tang, Jialong and Lin, Huan and Yang, Baosong and Xie, Pengjun and Huang, Fei and others},
	year         = 2024,
	journal      = {arXiv preprint arXiv:2407.19669}
}

@article{wang2024multilingual,
	title        = {Multilingual e5 text embeddings: A technical report},
	author       = {Wang, Liang and Yang, Nan and Huang, Xiaolong and Yang, Linjun and Majumder, Rangan and Wei, Furu},
	year         = 2024,
	journal      = {arXiv preprint arXiv:2402.05672}
}

@article{xu2023demystifying,
	title        = {Demystifying clip data},
	author       = {Xu, Hu and Xie, Saining and Tan, Xiaoqing Ellen and Huang, Po-Yao and Howes, Russell and Sharma, Vasu and Li, Shang-Wen and Ghosh, Gargi and Zettlemoyer, Luke and Feichtenhofer, Christoph},
	year         = 2023,
	journal      = {arXiv preprint arXiv:2309.16671}
}

@incollection{fisher1970statistical,
	title        = {Statistical methods for research workers},
	author       = {Fisher, Ronald Aylmer},
	year         = 1970,
	booktitle    = {Breakthroughs in statistics: Methodology and distribution},
	publisher    = {Springer},
	pages        = {66--70}
}

@article{agrawal2024shelfhelp,
	title        = {ShelfHelp: Empowering humans to perform vision-independent manipulation tasks with a socially assistive robotic cane},
	author       = {Agrawal, Shivendra and Nayak, Suresh and Naik, Ashutosh and Hayes, Bradley},
	year         = 2024,
	journal      = {arXiv preprint arXiv:2405.20501}
}

%%%%%%%%%%%%%%%%%%%%%%%%%%%%%%%%%%%%%%%%%%%%%%%%%%%%%%%%%%%%%%%%%%%%%%%%

\end{document}